\documentclass[]{spie}  

\usepackage{amssymb}
\usepackage{pifont}

\usepackage{amsmath,amsfonts,amssymb}
\usepackage{subcaption}
\usepackage{graphicx}
\usepackage{cite} 
\usepackage{times}
\usepackage{epsfig}
\usepackage{amsmath}
\usepackage{nccmath}
\usepackage{amssymb}
\usepackage{mwe}
\usepackage{acro}
\usepackage{amssymb}
\usepackage{xcolor,colortbl}
\usepackage{tabularx}
\usepackage{relsize}
\usepackage{pifont}
\usepackage{booktabs} 
\usepackage{multirow}
\usepackage{multicol}
\usepackage{adjustbox}
\usepackage{float}
\usepackage{graphicx}
\usepackage{makecell}
\usepackage{tabu}
\usepackage[colorlinks=true, allcolors=blue]{hyperref}
\usepackage[capitalize]{cleveref}

\title{DyMorph-B2I: Dynamic and Morphology-Guided Binary-to-Instance Segmentation for Renal Pathology}

\author[a]{Leiyue Zhao}
\author[b]{Yuechen Yang}
\author[b]{Yanfan Zhu}
\author[c]{Haichun Yang}
\author[b,c]{Yuankai Huo}
\author[d]{Paul D. Simonson}
\author[d]{Kenji Ikemura}
\author[d,e]{Mert R. Sabuncu}
\author[d,f]{Yihe Yang}
\author[b,d]{Ruining Deng}

\affil[a]{Southern University of Science and Technology, Shenzhen, Guangdong, 518055, CN}
\affil[b]{Vanderbilt University, Nashville, TN, 37235, USA}
\affil[c]{Vanderbilt University Medical Center, Nashville TN 37232, USA}
\affil[d]{Weill Cornell Medicine, New York, NY 10065, USA}
\affil[e]{Cornell Tech, New York, NY 10044, USA}
\affil[f]{Northwell Health, New Hyde Park, NY 11040, USA}

\begin{document} 
\maketitle

\begin{abstract}
Accurate morphological quantification of renal pathology functional units relies on instance-level segmentation, yet most existing datasets and automated methods provide only binary (semantic) masks, limiting the precision of downstream analyses. Although classical post-processing techniques such as watershed, morphological operations, and skeletonization, are often used to separate semantic masks into instances, their individual effectiveness is constrained by the diverse morphologies and complex connectivity found in renal tissue. In this study, we present DyMorph-B2I, a dynamic, morphology-guided binary-to-instance segmentation pipeline tailored for renal pathology. Our approach integrates watershed, skeletonization, and morphological operations within a unified framework, complemented by adaptive geometric refinement and customizable hyperparameter tuning for each class of functional unit. Through systematic parameter optimization, DyMorph-B2I robustly separates adherent and heterogeneous structures present in binary masks. Experimental results demonstrate that our method outperforms individual classical approaches and naïve combinations, enabling superior instance separation and facilitating more accurate morphometric analysis in renal pathology workflows. The pipeline is publicly available at: https://github.com/ddrrnn123/DyMorph-B2I.

\end{abstract}


\section{Description of purpose}
\label{sec:intro} 
Accurate morphological quantification of renal pathology functional units depends on the ability to distinguish individual structures through instance-level segmentation~\cite{8099955,zhu2025cross,ke2024tshfna}. Yet, most existing datasets and automated tools provide only binary (semantic) masks~\cite{deng2023omni,deng2024prpseg,deng2025segment,deng2025casc,deng2023democratizing,jayapandian2021development,bouteldja2021deep}, in which all pixels of a given class are labeled collectively, without regard for the identity or boundaries of individual anatomical entities. While semantic segmentation facilitates basic assessments of tissue composition, it falls short for quantitative morphometric analysis, where precise object-level measurements such as the size, shape, or spatial distribution of glomeruli, tubules, and other functional units are essential~\cite{JIANG20211431,yang2025pyspatial,chen2024spatial}. Critically, semantic masks are unable to resolve adjacent or overlapping structures and cannot provide reliable counts or object-specific statistics. These limitations pose significant challenges for downstream analyses, including the evaluation of pathological heterogeneity~\cite{carmichael2022incorporatingintratumoralheterogeneityweaklysupervised,zhu2025cross}, detection of rare subtypes~\cite{Quellec_2020,yu2025glo}, and the development of advanced models that rely on instance-level annotations for accuracy and interpretability~\cite{Jiang2021Glomeruli}. As a result, the absence of robust instance segmentation methods remains a major obstacle to advancing computational pathology in both research and clinical settings.

\begin{figure}[htbp]            
  \centering
  \resizebox{0.8\textwidth}{!}{\includegraphics{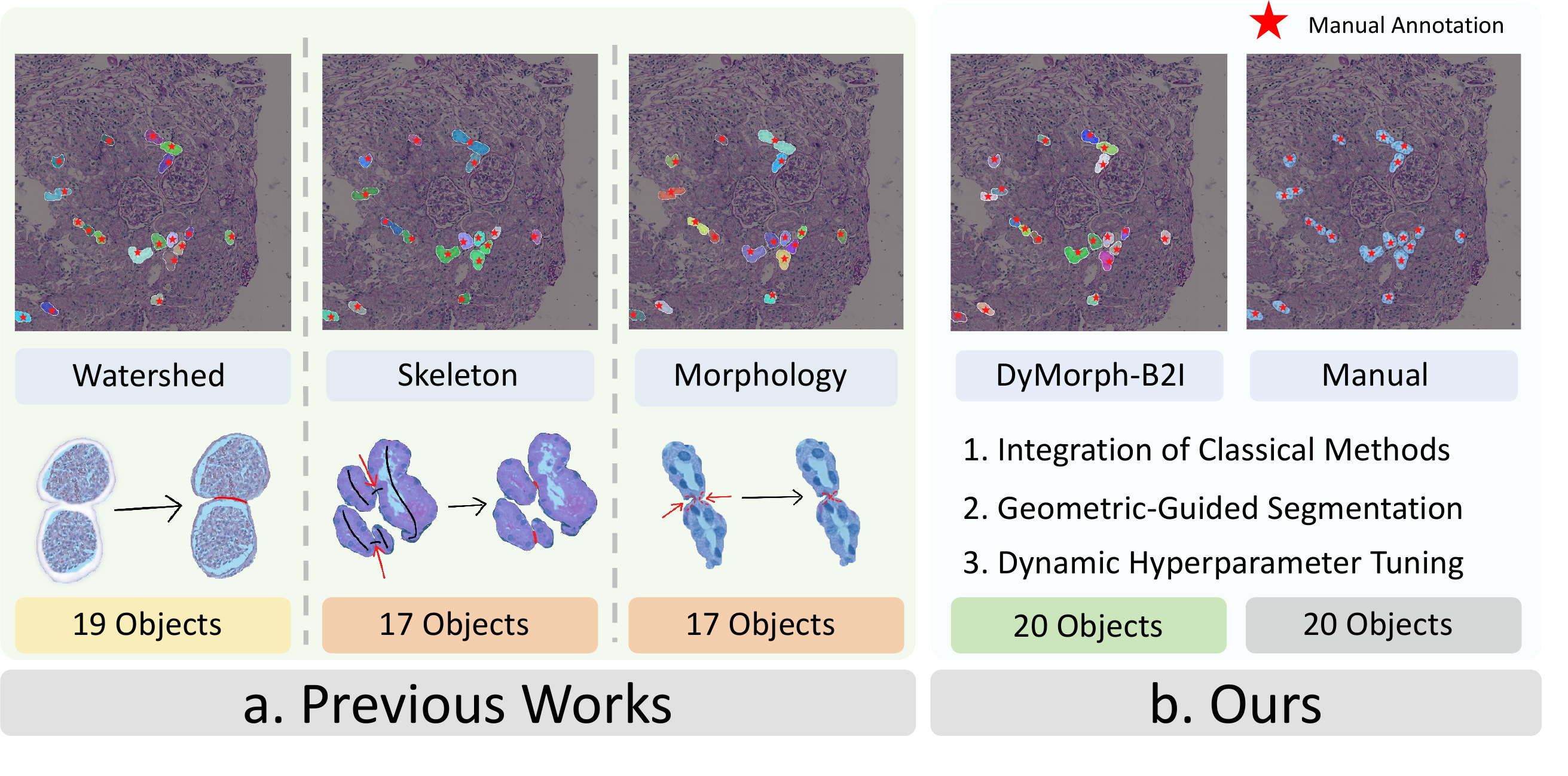}}
  \caption{Comparison of instance count among traditional methods, DyMorph-B2I, and manual annotation. The same semantic mask is processed by different methods to generate instance segmentation results, with an explanation of the traditional approach included. Numbers indicate the detected instance counts for each method. }
  \label{fig:compare}
\end{figure}

Classical post-processing techniques such as watershed~\cite{RAMBABU2007210}, morphological operations~\cite{4767941}, and skeletonization~\cite{Zhang1984} are commonly used to convert semantic masks into instance-level results. However, their effectiveness is often limited by the complex and heterogeneous nature of renal tissue. Each algorithm brings its own strengths, yet none can fully address the wide variability in shapes and connectivity observed across different functional units. Consequently, these traditional methods frequently struggle to separate connected or overlapping instances, particularly in regions characterized by intricate morphologies or tightly packed structures, as shown in Fig.~\ref{fig:compare}.

To overcome these challenges, we propose DyMorph-B2I,a dynamic, morphology-guided binary-to-instance segmentation pipeline tailored for renal pathology. Our method integrates watershed, skeletonization, and morphological operations within a unified framework, enhanced by adaptive geometric refinement and customizable hyperparameter tuning that accommodates the distinct morphological features of each functional unit. By dynamically optimizing these parameters for each image, DyMorph-B2I achieves robust separation of adherent structures in binary masks, irrespective of their underlying complexity.

Through both qualitative and quantitative evaluation, we demonstrate that DyMorph-B2I surpasses the performance of individual classical methods and naïve combinations, delivering superior instance separation across diverse renal tissue types. This flexible pipeline not only streamlines data preparation for downstream instance segmentation tasks but also enables more accurate and comprehensive analyses in renal pathology.

\section{Methods}

\begin{figure}[htbp]            
  \centering
  \resizebox{0.9\textwidth}{!}{\includegraphics{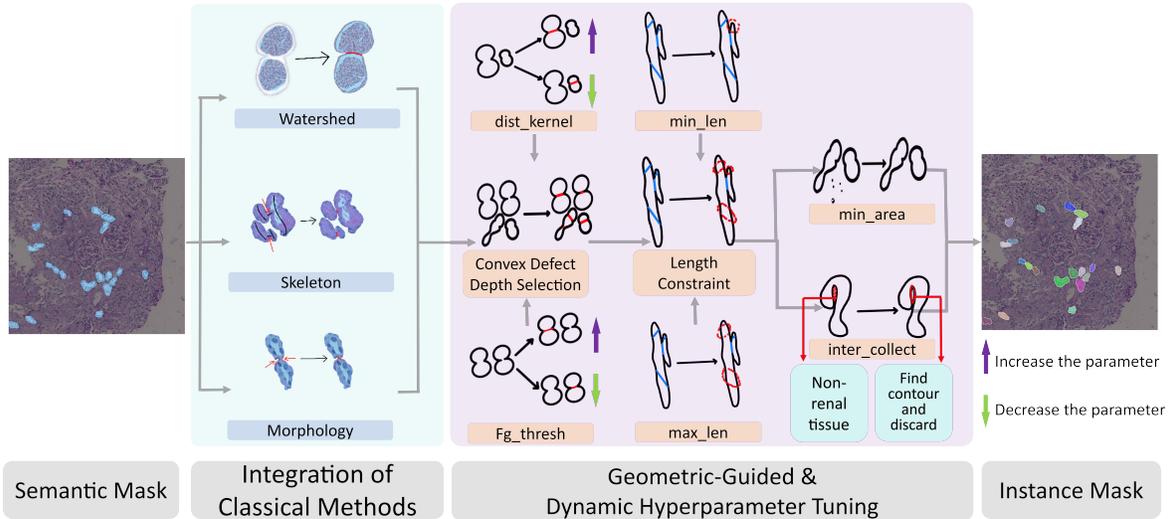}}
  \caption{The workflow of the DyMorph-B2I method. The intermediate visualizations illustrate the function and effect of each processing step. The process begins with the integration of traditional image processing techniques (including watershed, skeletonization, and morphological operations) to generate initial instance candidates. This is followed by a geometric-guided dynamic hyperparameter tuning module that further refines the results, including steps such as convex defect depth selection, length constraint, minimum area filtering and optional use of inner contour extraction. The final output is an instance-level segmentation mask. }
  \label{fig:methods}
\end{figure}

\subsection{Binary-to-Instance Pipeline with Integration of Classical Methods}
We present an integrated segmentation pipeline that incorporates three fundamental image processing techniques: watershed, skeletonization, and mathematical morphology. Each method offers distinct advantages but also exhibits specific limitations when applied to renal pathology. The schematic principles underlying these traditional approaches are illustrated in the Integration of Classical Methods section of Figure.~\ref{fig:methods}.

\begin{enumerate}
\item \textbf{Watershed:} Drawing on topological principles as outlined by Rambabu et al.~\cite{RAMBABU2007210}, the watershed algorithm conceptualizes images as topographic surfaces, delineating regions according to local minima and their corresponding watershed lines. This approach is particularly effective for isolating circular anatomical features, such as renal caps and tufts. Nevertheless, it tends to underperform when confronted with elongated or tightly clustered structures, often resulting in under-segmentation.

\item \textbf{Skeleton:} Skeletonization, following the methodology of Lee et al.~\cite{Zhang1984}, reduces binary regions to their central skeletal forms, thereby facilitating the segmentation of elongated features by highlighting subtle branches connecting larger domains. However, its utility diminishes with more compact or rounded objects, and it may fragment structures within densely packed regions.

\item \textbf{Morphology:} Morphological operations~\cite{4767941}, specifically erosion and dilation, are adept at introducing separation points to resolve minor adhesions and clarify object boundaries. While well suited for disentangling lightly connected regions, these methods often struggle in the face of significant morphological diversity or densely populated areas, where variability in object size and shape can complicate segmentation.
\end{enumerate}

Given the complex composition of renal pathology images, which often feature an interplay of round, elongated, and variably connected structures, our pipeline strategically leverages these complementary algorithms. The process begins with the application of the watershed algorithm to each class-specific binary segmentation mask, targeting the preliminary separation of circular adhesions. Subsequently, convexity defect analysis is employed to scrutinize larger or elongated regions, enabling the detection of concave boundaries that may indicate true segmentation interfaces. Finally, skeletonization is integrated to further refine or augment segmentation lines, thereby enhancing boundary precision and facilitating a more accurate partitioning of individual structures into distinct instances.

\subsection{Geometric-Guided Segmentation Line Refinement}

While a simple combination of the three classical methods can, at times, yield higher error rates than certain traditional approaches, our primary objective is to ensure the accurate delineation of adherent regions. This approach emphasizes the generation of plausible segmentation lines, even if it occasionally produces superfluous boundaries. When this integrated strategy is paired with a subsequent fine-tuning step, the pipeline achieves a marked improvement in segmentation performance.

To address the challenges posed by complex tissue morphologies, we implement a morphology-aware guidance algorithm that systematically refines candidate segmentation lines based on geometric context. This refinement process is structured around two key components:

\begin{enumerate}
\item \textbf{Convex Defect Depth Selection and Skeleton Alignment Correction.}
Following convexity defect analysis, all defect points and their respective depths are identified. We select the two deepest defects and use their coordinates as provisional endpoints for candidate segmentation lines. These preliminary lines are then compared to the skeleton of the target region, which approximates the true medial axis of the anatomical structure, as illustrated in the Convex Defect
Depth Selection section of Figure.~\ref{fig:methods}. If a candidate line exhibits substantial deviation from the nearest skeleton segment, its position is adjusted to achieve closer alignment, thereby ensuring that the final segmentation line more faithfully represents the true boundary.

\item \textbf{Length Constraint.}
In the Length Constraint section of Figure.~\ref{fig:methods}, each candidate line is subjected to a length constraint, calculated as the Euclidean distance between its endpoints. To ensure anatomical plausibility, we adaptively determine the acceptable range of lengths using a bisection method, which iteratively refines the interval to exclude outliers and converge upon suitable bounds for segmentation.
\end{enumerate}

\subsection{Dynamic and Customizable Hyperparameter Tuning}

To optimize our segmentation pipeline for diverse image contexts, we implemented a dynamic hyperparameter tuning strategy that adapts to the unique characteristics of each image. Hyperparameter tuning was guided by expert anatomical knowledge, with the object count serving as a primary reference for evaluating plausibility. This approach enables precise and efficient instance separation, as illustrated in the Geometric-Guided \&
Dynamic Hyperparameter Tuning section of Figure.~\ref{fig:methods}, by systematically exploring key algorithmic parameters:

\begin{enumerate}
    \item \textbf{mix\_params}: We define this as a set of hyperparameter tuples, allowing for the parallel evaluation of multiple segmentation configurations via grid search or comparative testing. Each tuple encompasses five essential elements:
    \begin{itemize}
        \item \textbf{method}: Specifies the segmentation technique or combination of methods applied. By incorporating both classical and hybrid approaches, this parameter broadens the search for optimal cut line precision.
        \item \textbf{dist\_kernel}: Controls the size of the distance-transform kernel. Increasing the kernel size can help separate closely packed objects, though it may overlook slender or small targets. Conversely, a smaller kernel retains finer edge detail but may result in over-segmentation or excess foreground seeds.
        \item \textbf{fg\_thresh}: Sets the foreground-threshold ratio, influencing how adjacent objects are partitioned. Higher values enhance object separation but risk over-segmentation and the loss of small targets, whereas lower thresholds encourage conservative splitting and may merge distinct regions.
        \item \textbf{min\_len}: Establishes a lower bound for segmentation-line length, eliminating short segments that are likely to represent noise or insignificant micro-cuts.
        \item \textbf{max\_len}: Places an upper limit on segmentation-line length, preventing the formation of overly long lines that could inadvertently divide whole regions.
    \end{itemize}

    \item \textbf{min\_area}: Sets a minimum area requirement for contours. By filtering out small, spurious objects and noise, this parameter ensures that only meaningful structures are included in the final segmentation.

    \item \textbf{inter\_collect}: A Boolean flag indicating whether internal, nested contours should be extracted. This option is particularly valuable for accurately capturing structures with internal holes or for recovering regions missing from binary masks.
\end{enumerate}

\section{Data \& Experiments}
\subsection{Data}

The dataset for this study was curated from the NEPTUNE project~\cite{barisoni2013digital}, comprising multiple renal tissue components, including distal tubules (DT), proximal tubules (PT), glomerular capsules (CAP), glomerular tufts (TUFT), arteries (ART), and peritubular capillaries (PTC). Expert annotators manually delineated regions of interest (ROIs) to capture four distinct morphological object types exhibiting normal histological architecture. Each ROI was acquired at a resolution of \( 3000 \times 3000 \) pixels, corresponding to \( 40\times \) magnification with a pixel size of \( 0.25\, \mu\text{m} \). Images represent a range of staining modalities, including Hematoxylin and Eosin (H\&E), PAS, Silver Stain (SIL), and Trichrome Stain (TRI). For each class, 50 ROIs were selected, with a preference for more challenging images. In total, the dataset comprises 300 images, providing a diverse and representative sample for downstream analysis.

\subsection{Evaluation Metrics}
To ensure a robust and meaningful comparison of segmentation methods, we introduce two dedicated evaluation metrics.

The first metric, Mean Absolute Percentage Error (MAPE), quantifies the average magnitude of error relative to the ground truth and is calculated as follows:
\begin{equation}
\mathrm{MAPE}_{c}
= \frac{1}{\lvert S_{c}\rvert}
\sum_{i \in S_{c}}
\frac{\bigl\lvert m_{i} - h_{i}\bigr\rvert}{h_{i}},
\label{eq:mape}
\end{equation}
\noindent Where $m_{i}$ denotes the measurement produced by the segmentation method for sample $i$, $h_{i}$ represents the corresponding human (ground-truth) value, and $\lvert S_{c}\rvert$ is the total number of samples in class $c$.

The second metric, Percentage Error (PE), is formulated as
\begin{equation}
\mathrm{PE}_{c}
=
\frac{m_i - h_i}{h_i},
\label{eq:pe}
\end{equation}

\noindent Using the notation introduced above, we distinguish between two complementary error metrics. Unlike MAPE, which measures the absolute discrepancy, PE captures the directionality of the error, indicating whether a method systematically overestimates or underestimates instance counts compared to the human reference.

We employ MAPE to quantitatively assess the segmentation accuracy of each method, where lower values reflect closer alignment with expert annotations. In parallel, PE serves as a qualitative indicator, revealing each method’s intrinsic bias: positive values suggest a tendency toward over-segmentation, whereas negative values indicate under-segmentation. Together, these complementary metrics provide a comprehensive evaluation framework for comparing segmentation performance across diverse morphological classes.

\begin{table*}[hbp]

\centering
\caption{
Error statistics per class and overall method-level statistics}

\resizebox{0.78\textwidth}{!}{
\begin{tabular}{l|ccccccccc}
\toprule
\multirow{2}{0.8in}{Method} & \multicolumn{3}{c}{DT } & \multicolumn{3}{c}{PT } & \multicolumn{3}{c}{CAP}\\
\cmidrule(lr){2-4}
\cmidrule(lr){5-7}
\cmidrule(lr){8-10}
 & Error$\downarrow$ & Spearman$\uparrow$ & Pearson$\uparrow$ & Error$\downarrow$ & Spearman$\uparrow$ & Pearson$\uparrow$ & Error$\downarrow$ & Spearman$\uparrow$ & Pearson$\uparrow$ \\
\midrule
FindContour ~\cite{SUZUKI198532}
  & 0.1486 & 0.8633 & 0.8304 
  & 0.1378 & 0.8654 & 0.8834 
  & 0.1395 & 0.8868 & 0.9220 \\

Watershed~\cite{RAMBABU2007210} 
  & 0.1337 & 0.9560 & 0.9324 
  & 0.1196 & 0.9417 & 0.9448 
  & 0.0588 & 0.9617 & 0.9561 \\

Skeleton~\cite{Zhang1984} 
  & 0.6366 & 0.6921 & 0.5375 
  & 2.4481 & 0.3312 & 0.2970 
  & 0.2647 & 0.8546 & 0.8766 \\

Morphology ~\cite{4767941}
  & 0.1369 & 0.9502 & 0.9334 
  & 0.1471 & 0.7984 & 0.8408 
  & 0.1562 & 0.8762 & 0.9178 \\

Combination 
  & 0.6534 & 0.9495 & 0.9364 
  & 0.6400 & 0.8598 & 0.8851 
  & 0.7750 & 0.8573 & 0.9189 \\

\midrule
DyMorph-B2I: (Ours) 
  & \textbf{0.0025} & \textbf{0.9974} & \textbf{0.9992} 
  & \textbf{0.0072} & \textbf{0.9875} & \textbf{0.9922} 
  & \textbf{0.0000} & \textbf{1.0000} & \textbf{1.0000} \\
\bottomrule

\end{tabular}
}

\resizebox{0.98\textwidth}{!}{
\begin{tabular}{l|cccccccccccc}
\toprule
\multirow{2}{0.8in}{Method} & \multicolumn{3}{c}{TUFT } & \multicolumn{3}{c}{ART } & \multicolumn{3}{c}{PTC } & \multicolumn{3}{c}{Average }\\
\cmidrule(lr){2-4}
\cmidrule(lr){5-7}
\cmidrule(lr){8-10}
\cmidrule(lr){11-13}
 & Error$\downarrow$ & Spearman$\uparrow$ & Pearson$\uparrow$ & Error$\downarrow$ & Spearman$\uparrow$ & Pearson$\uparrow$ & Error$\downarrow$ & Spearman$\uparrow$ & Pearson$\uparrow$ & Error$\downarrow$ & Spearman$\uparrow$ & Pearson$\uparrow$\\
\midrule
FindContour ~\cite{SUZUKI198532}
  & 0.0282 & 0.9863 & 0.9785 
  & 0.2700 & 0.9832 & 0.9875 
  & 0.0000 & 1.0000 & 1.0000
  & 0.1207 & 0.9308 & 0.9336 \\

Watershed~\cite{RAMBABU2007210} 
  & 0.0090 & 0.9432 & 0.9910 
  & 0.3060 & 0.9985 & 0.8423 
  & 0.1482 & 0.9201 & 0.9578 
  & 0.1292 & 0.9366 & 0.9374 \\

Skeleton~\cite{Zhang1984} 
  & 0.1365 & 0.9136 & 0.9194 
  & 0.2573 & 0.9024 & 0.9215 
  & 9.3812 & 0.5576 & 0.6603  
  & 2.1874 & 0.7086 & 0.7020 \\

Morphology ~\cite{4767941}
  & 0.0282 & 0.9863 & 0.9785 
  & 0.2600 & 0.9785 & 0.9857 
  & 0.4902 & 0.6956 & 0.7816  
  & 0.2031 & 0.8809 & 0.9063 \\

Combination 
  & 0.9162 & 0.9916 & 0.9665 
  & 0.4262 & 0.9665 & 0.9268 
  & 0.0770 & 0.9518 & 0.9926  
  & 0.5821 & 0.9294 & 0.9377 \\
\midrule
DyMorph-B2I: (Ours) 
  & \textbf{0.0000} & \textbf{1.0000} & \textbf{1.0000} 
  & \textbf{0.0000} & \textbf{1.0000} & \textbf{1.0000} 
  & \textbf{0.0000} & \textbf{1.0000} & \textbf{1.0000}  
  & \textbf{0.0016} & \textbf{0.9975} & \textbf{0.9986}\\
\bottomrule
\end{tabular}
}
\label{table:result}
\end{table*}

\section{Results}

\subsection{Instance Segmentation Performance}

\begin{figure}[htbp]            
  \centering
  \resizebox{0.8\textwidth}{!}{\includegraphics{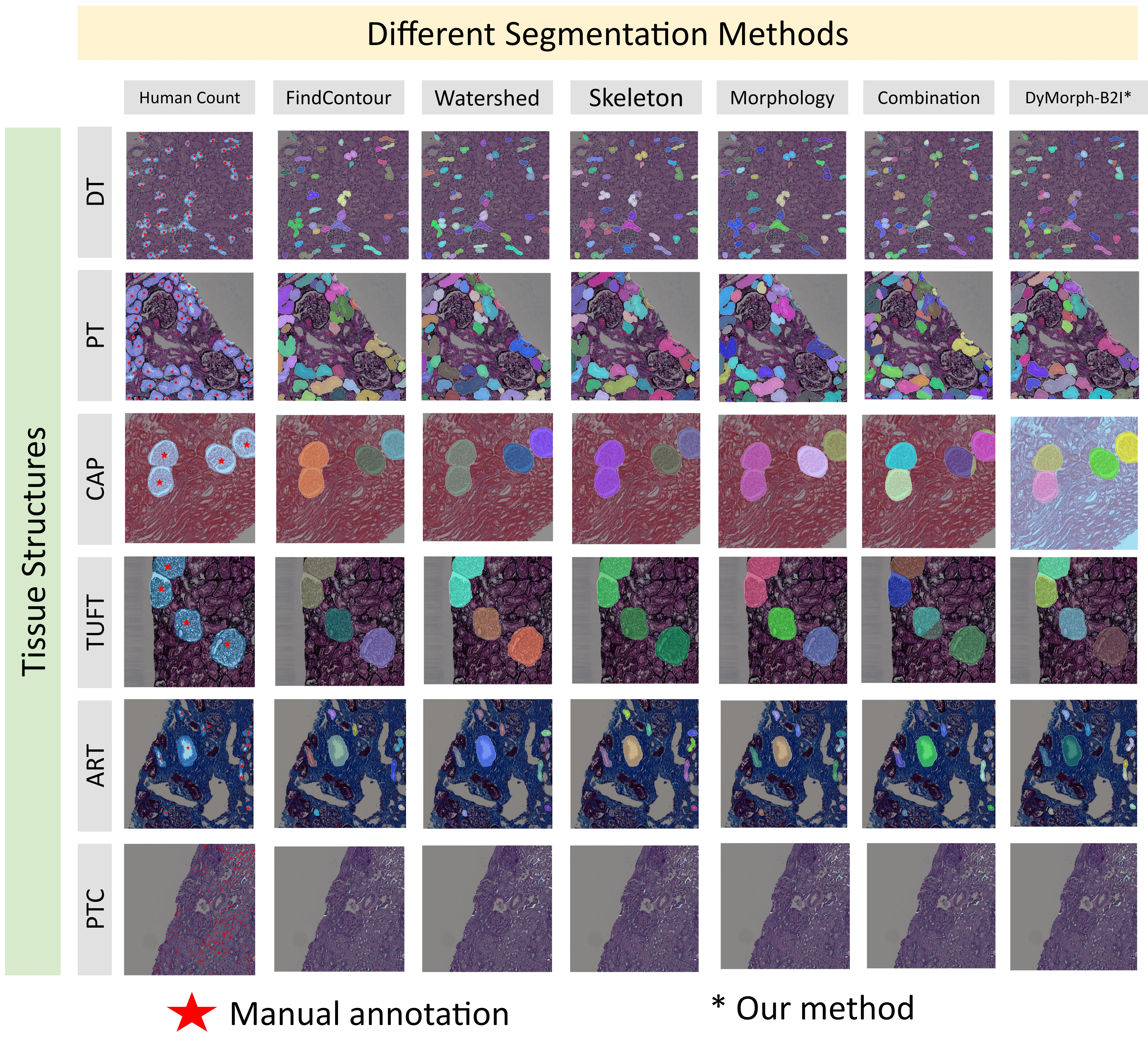}}
  \caption{Comparison of segmentation performance across six classes using classical methods, the combination approach, and DyMorph-B2I. The proposed method demonstrates superior instance segmentation performance when converting semantic masks. Evaluation was based on the object number determined by clinical anatomy knowledge, i.e., the manual annotation shown in the figure.}
  \label{fig:eval}
\end{figure}

\begin{figure}[htbp]            
  \centering
  \resizebox{1\textwidth}{!}{\includegraphics{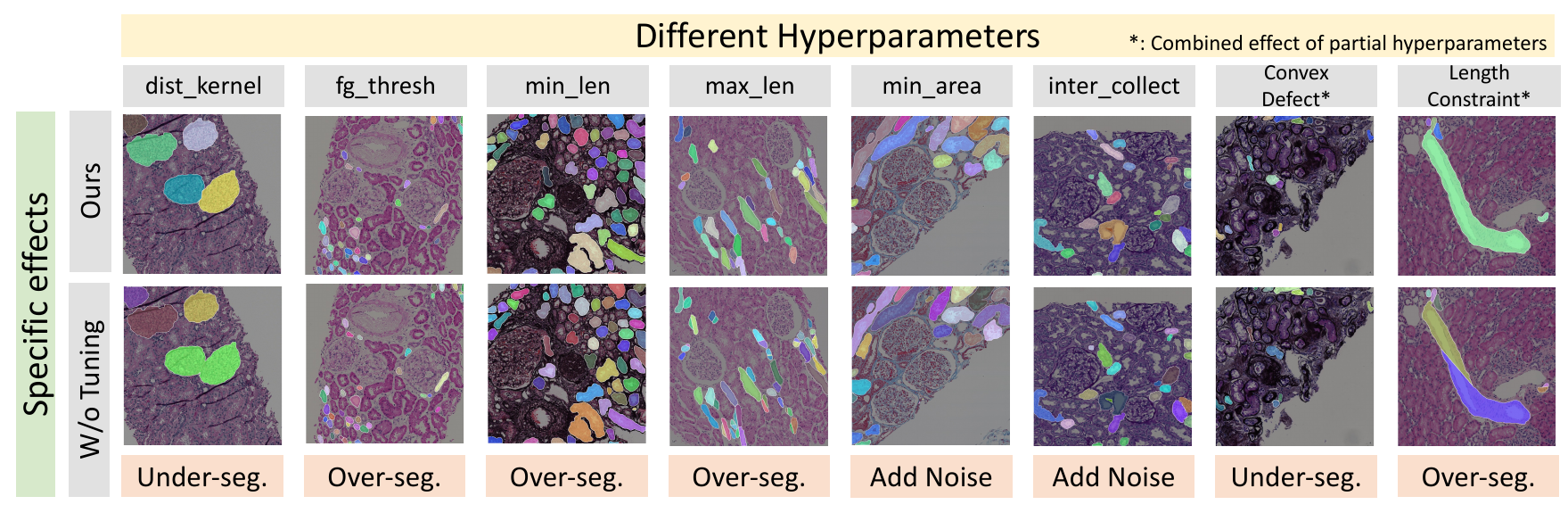}}
  \caption{Comparative analysis of segmentation results with and without specific hyperparameter tuning. The figure reveal the performance impact of each adjustment and underscore the adverse effects associated with the `` W/O Tunning".}
  \label{fig:parameter}
\end{figure}

Fig.\ref{fig:eval} and Table\ref{table:result} compare the performance of different instance separation methods across six renal pathology classes using MAPE. Traditional algorithms show clear limitations: FindContour performs well for well-defined structures such as PTC and TUFT, but struggles with complex morphologies like DT and PT; Watershed and Morphology excel in several classes yet lose accuracy with intricate shapes; and Skeletonization remains inconsistent, particularly for elongated or densely connected structures. A naïve Combination offers modest gains but introduces instability. In contrast, DyMorph-B2I consistently achieves high accuracy across all classes by integrating classical methods with morphology-aware refinement and adaptive hyperparameter tuning, demonstrating that tailored, adaptive strategies outperform uniform traditional approaches for reliable instance separation in renal pathology.

Fig.~\ref{fig:parameter} highlights the pivotal role of dynamic, customizable hyperparameter tuning in achieving precise instance segmentation. Without targeted adjustments, results suffer from under- or over-segmentation, noise amplification, and boundary misalignment, particularly in complex or adherent structures. In contrast, our tuned configuration adapts to morphological heterogeneity, yielding cleaner separations and more accurate structural representations.

\begin{figure}[htbp]            
  \centering
  \resizebox{1\textwidth}{!}{\includegraphics{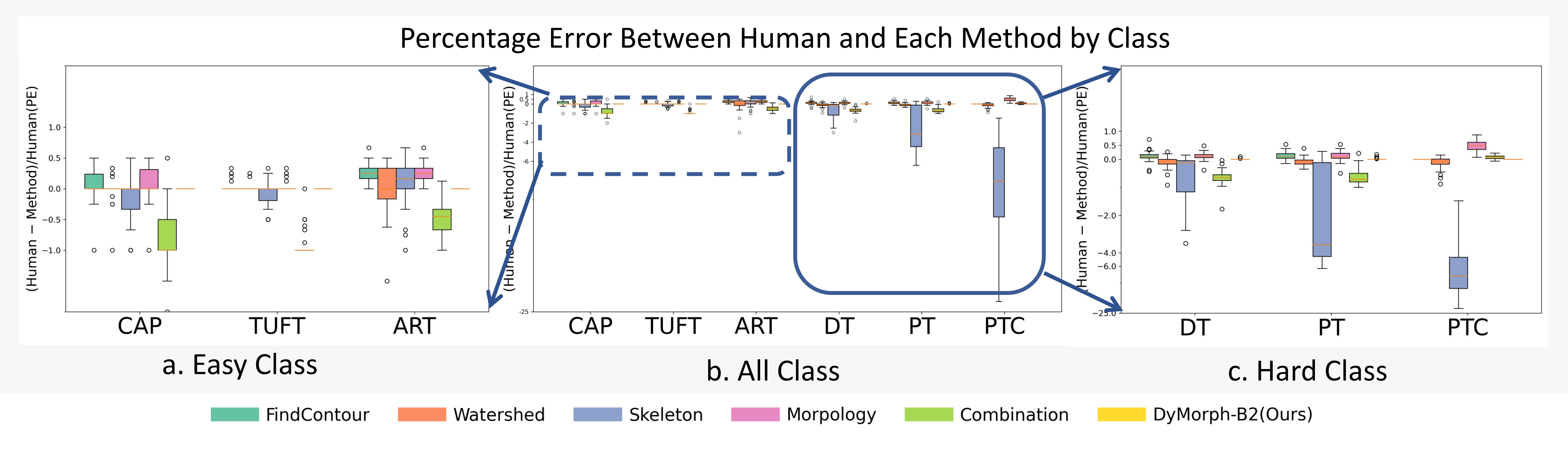}}
  \caption{Boxplots of PE for six segmentation algorithms.Panels show three scenarios: (a) Easy Class, in which all methods achieve performance levels comparable to human annotations; (b) All Class, representing the combined results of all six categories, with left inset zooming into the low‐error range and right inset highlighting larger deviations; and (c) Hard Class, representing the most challenging cases, where errors are both larger in magnitude and more variable. Positive values indicate under-segmentation by the method, while negative values indicate over-segmentation.}
  \label{fig:boxplot}
\end{figure}

Fig.~\ref{fig:boxplot} presents the quantitative results with metrics PE, revealing clear contrasts in performance across difficulty levels. For relatively simple categories such as CAP, TUFT, and ART, most methods maintain consistently low error rates. In morphologically complex classes like DT, PT, and PTC, however, accuracy declines sharply, with errors showing both increased variance and systematic bias. These reductions in performance are largely attributable to the difficulty of segmenting elongated or densely connected structures, which often results in pronounced over-segmentation. This weakness is particularly apparent in the Skeletonization method, where error distributions are both broader and less stable. In contrast, the tuned configuration yields a more compact error distribution, indicating that targeted hyperparameter optimization can substantially enhance stability while reducing extreme deviations.

\subsection{Specific Structure Categories}

By systematically analyzing each structure category, distinct strengths and limitations of individual segmentation methods become apparent. 

For \textbf{CAP} structures, which exhibit regular circular or hemispherical contours with moderate grayscale contrast, FindContour and Watershed reliably delineate boundaries, while Morphology also performs well. However, Skeleton frequently misclassifies small skeletal branches as segmentation lines, leading to unnecessary splits. When combination methods are applied, inconsistencies between component outputs, such as mismatches between morphological closing results and watershed boundaries, frequently produce misaligned or inaccurate contours.

The segmentation of \textbf{DT} and \textbf{PT} introduces greater complexity due to their irregular and elongated shapes, non-uniform widths, and frequent adjacency to other features. Here, FindContour’s reliance on uninterrupted contour extraction often causes it to miss or over-detect discontinuous edges. Skeletonization, intended to abstract centerlines, proves highly sensitive to image noise, which can fragment or distort the medial axis. Both Watershed and Morphology, despite their strengths in less complex regions, tend to over-expand along structural cracks, occasionally merging distinct entities and thereby increasing segmentation errors. Notably, for PT, Watershed’s tendency to over-segment dense clusters and Morphology’s propensity to merge adjacent particles both contribute to systematic under- or overestimation of object counts, while Skeletonization fails to resolve individual particles altogether.

For \textbf{PTC}, which are exceptionally small and can be distributed sparsely or densely, the performance of all methods deteriorates. While FindContour maintains accuracy for isolated points due to its sensitivity to intensity changes, Watershed at low thresholds can misclassify background noise, and Morphology struggles to select structuring elements that differentiate true PTC from artifacts. Skeletonization is largely ineffective for such minimal structures. Furthermore, when errors from individual methods are compounded in naïve combinations, stable extraction of these features becomes even more elusive.

\textbf{TUFT} and \textbf{ART} categories present their own challenges. Although TUFT are relatively regular and circular, their dense packing leads Skeletonization to generate spurious junctions, while Morphology is unable to resolve heavy adhesion, often causing fragmentation or the loss of fine detail. ART structures, characterized by elongated, thread-like forms and jagged edges, pose a significant challenge to Watershed, which is prone to fragmentation under noisy conditions. Morphological methods are limited by the substantial adhesion between ART structures, and Skeletonization often interprets jagged boundaries as extraneous branches, further elevating segmentation errors.

Taken together, these observations highlight the intrinsic limitations of relying on a single classical method to address the wide morphological diversity found in renal tissue. By comparison, the tuning-combination strategy adaptively integrates the strengths of multiple techniques while dynamically refining boundaries, enabling it to better align segmentation outputs with the specific characteristics of each structural category. This comprehensive approach results in more robust, reliable, and accurate performance across the full spectrum of renal pathology targets.

\subsection{Discussion and Future Work}

Although DyMorph-B2I demonstrates robust performance across diverse renal pathology structures, several limitations warrant consideration. Segmentation accuracy may be reduced in challenging cases such as DT and PT, where severe adhesion and complex morphologies can hinder complete object separation. Moreover, while the dynamic tuning and refinement steps enhance adaptability, they also increase computational load and extend processing times, especially for large or highly intricate datasets. Despite these constraints, DyMorph-B2I offers a transparent and effective solution for binary-to-instance conversion. Future work will explore the integration of advanced learning-based methods to further improve segmentation efficiency and accuracy.

\section{new or breakthrough work to be presented}

DyMorph-B2I is a dynamic, morphology-guided framework that integrates watershed, morphological, and skeleton-based methods with adaptive, image-specific tuning to transform semantic masks into precise instance segmentations. This tailored approach addresses structural heterogeneity and complex interconnections, achieving high concordance with expert annotations and providing a robust foundation for advanced morphological analysis in computational pathology.

\section{Conclusion}

We introduced DyMorph-B2I, a dynamic, morphology-guided pipeline that integrates classical image processing with adaptive refinement to achieve precise separation of complex and adherent structures in renal pathology. Although segmentation accuracy can decline for the most intricate morphologies, particularly in cases involving severe adhesion, and adaptive tuning may increase computational requirements, these trade-offs are balanced by the method’s robustness and adaptability. This framework establishes a strong platform for advancing instance segmentation in computational pathology, enabling more accurate morphological quantification, deeper characterization of pathological heterogeneity, and improved detection of rare subtypes in future studies.

\section{ACKNOWLEDGMENTS} 
This research was supported by the WCM Radiology AIMI Fellowship, WCM CTSC 2026 Pilot Award, NIH R01DK135597 (Huo), DoD HT9425-23-1-0003 (HCY), and the KPMP Glue Grant.

\bibliography{main} 
\bibliographystyle{spiebib} 

\end{document}